\documentclass{llncs}
\usepackage[utf8]{inputenc}
\usepackage{graphicx}
\usepackage{amsmath}
\usepackage{amssymb}
\usepackage{amsfonts}
\usepackage{amstext}
\usepackage{booktabs}
\usepackage{multirow}
\usepackage[table,xcdraw]{xcolor}
\usepackage{footmisc}
\newcommand{\avsum}{\mathop{\mathpalette\avsuminner\relax}\displaylimits}

\makeatletter
\newcommand\avsuminner[2]{%
  {\sbox0{$\m@th#1\sum$}%
   \vphantom{\usebox0}%
   \ooalign{%
     \hidewidth
     \smash{\vrule height\dimexpr\ht0+1pt\relax depth\dimexpr\dp0+1pt\relax}%
     \hidewidth\cr
     $\m@th#1\sum$\cr
   }%
  }%
}
\makeatother

\title{UVA Resources for the Biomedical Vocabulary Alignment at Scale in the UMLS Metathesaurus} 
\author{Vinh Nguyen, Olivier Bodenreider}
\institute{National Library of Medicine, National Institute of Health, Maryland, USA}

\begin{document}

\maketitle
\begin{abstract}
The construction and maintenance process of the UMLS (Unified Medical Language System) Metathesaurus  is time-consuming, costly, and error-prone as it relies on (1) the lexical and semantic processing for suggesting synonymous terms, and (2) the expertise of UMLS editors for curating the suggestions. For improving the UMLS Metathesaurus construction process, our research group has defined a new task called UVA (UMLS Vocabulary Alignment) and generated a dataset for evaluating the task. Our group has also developed different baselines for this task using logical rules (RBA), and neural networks (LexLM and ConLM).

In this paper, we present a set of reusable and reproducible resources including (1) a dataset generator, (2) three datasets generated by using the generator, and (3) three baseline approaches. We describe the UVA dataset generator and its implementation generalized for any given UMLS release. We demonstrate the use of the dataset generator by generating datasets corresponding to three UMLS releases, 2020AA, 2021AA and 2021AB. We provide three UVA baselines using the three existing approaches (LexLM, ConLM, and RBA). The code, the datasets, and the experiments are publicly available, reusable, and reproducible with any UMLS release (a no-cost license agreement is required for downloading the UMLS). 

\end{abstract}

\section{Introduction}
\label{introduction}

The Unified  Medical  Language  System  (UMLS) Metathesaurus is a biomedical terminology integration system developed by the US National Library of Medicine to integrate biomedical terms by organizing clusters of synonymous terms into concepts. The current construction process of the UMLS Metathesaurus heavily relies on lexical similarity algorithms to identify candidates for synonymy, although terms that do not share a common semantics are prevented from being recognized as synonymous. Additionally, synonymy asserted between terms in a source vocabulary is generally preserved in the Metathesaurus. Final synonymy determination comes from human curation by the Metathesaurus editors. Given the current size of the Metathesaurus with over 15 million terms from 214 source vocabularies, this process is inevitably costly and error-prone (as pointed out by \cite{cimino1998auditing,cimino2003consistency,jimeno2009reuse,morrey2009neighborhood,mougin2009analyzing}.) 

\textbf{Motivation.}
Recent successes in the field of natural language processing as well as the enormous knowledge accumulated over 30 years of manual curation has provided us ample material to tackle the problem of reconstructing the UMLS Metathesaurus using supervised learning approaches. We consider the reconstructing process as a large-scale synonymy prediction problem. The full scale of the this problem (e.g., $8.7M^2$ in the UMLS 2020AA version) is extremely large and impractical (even when restricted to English terms from active vocabularies), since $8.7M$ biomedical terms need to be compared pairwise. Therefore, our strategy for addressing this problem is finding the best performing approaches in the sampling evaluations as candidates for later full scale evaluations. We have recently achieved preliminary results \cite{bajaj2021evaluating,nguyen2021adding,nguyen2022conlm,nguyen2021biomedical,wijesiriwardene2022ubert} in the sampling evaluation phase. 
 
In our recent work \cite{nguyen2021biomedical}, we formalized synonymy prediction in the UMLS Metathesaurus as a vocabulary alignment problem, referred to as the UMLS Vocabulary Alignment (UVA) problem. Given a pair of biomedical terms from the UMLS source vocabularies, we developed different approaches for predicting if the terms are synonymous. For evaluating any approach for the UVA task, we generated a set of dataset variants (also referred to as ``seed alignment'' or ``ground truth'') with different degrees of lexical similarity among the negative examples as described in \cite{nguyen2021biomedical}. These UVA datasets containing hundreds of millions of pairs of biomedical terms from the active subset of source vocabularies in the UMLS 2020AA have been used for evaluating the approaches as follows.

Our recent work \cite{bajaj2021evaluating,nguyen2021adding,nguyen2022conlm,nguyen2021biomedical,wijesiriwardene2022ubert} developed several approaches for addressing the UVA problem. In \cite{nguyen2021biomedical}, we described the RBA (rule-based approximation) approach that approximates the current construction process into a set of rules. In \cite{nguyen2021biomedical}, we also described LexLM, a supervised learning approach that utilized the Siamese architecture with LSTM and BioWordVec \cite{zhang2019biowordvec} embeddings. Although the LexLM largely outperformed the RBA, we noted as a limitation of this work that we only leveraged lexical information and did not include any contextual information. In \cite{nguyen2022conlm}, we addressed this limitation of the LexLM model by incorporating the contextual information into the ConLM model. We also attempted to improve the performance of the LexLM by adding an attention layer into the neural network in \cite{nguyen2021adding}. In \cite{bajaj2021evaluating} we evaluated the role of different biomedical embeddings such as BioBERT \cite{lee2020biobert} and SapBERT \cite{liu2020self} compared to the BioWordVec in the LexLMs. Additionally, in \cite{wijesiriwardene2022ubert} we fine-tuned the BERT models such as BioBERT and SapBERT for the UVA task. 

\textbf{Shared resources.} 
We believe that sharing our recent work as reusable and reproducible resources will benefit broader research communities in many different directions. Besides using our work in the internal applications at the National Library of Medicine, we are interested in engaging and supporting external research communities. While exploring several techniques as described above, given the large scale and domain diversity of the UVA datasets compared to the related work, we believe that the UVA problem and its extended biomedical applications will offer various multi-disciplinary challenges for different research communities. Therefore, our goal in this paper is to prepare the materials suitable for the UVA challenge with three specific resources as follows. 

\textbf{Contributions.} We share the following resources with broader research communities:
\begin{itemize}
\item A scalable dataset generator taking input files from any UMLS version and generate UVA datasets with customizable parameters (Section \ref{generator}). The new UMLS releases will be used for generating the new UVA challenge datasets.
\item A set of UVA datasets from the UVA generator, particularly for the UMLS Metathesaurus 2020AA, 2021AA and 2021AB  (Section \ref{generating_datasets}).
\item A set of UVA baseline approaches (Section \ref{baselines}) and reproducible experiments including RBA, LexLM, and ConLM (Section \ref{experiments}).
\end{itemize}

These resources are connected in a multi-step pipeline from reading the input files from a particular UMLS Metathesaurus,  generating the datasets, and running the experiments with the baseline approaches. All the UVA resources including the source code, the datasets, and the instructions for running the experiments are available at \url{https://w3id.org/uva}. 

The remainder of the paper is organized as follows. Section \ref{background} provides background knowledge in the Metathesaurus. Section \ref{generator} provides the key information about the dataset generator. Section \ref{baselines} summarizes the three baseline approaches for the UVA task. In Section \ref{experiments}, we present our experiments for the three baselines and their results. Section \ref{discussion} discusses our future work and the maintenance plan for the UVA resources. In section \ref{related}, we discuss the related work. Section \ref{conclusion} concludes the paper.

\section{Knowledge Representation in the UMLS Metathesaurus for the UVA Problem}
\label{background}

This section summarizes the key concepts in the UMLS Metathesaurus to be used in the UVA problem. Additional details and examples can be found in \cite{nguyen2022conlm,nguyen2021biomedical} and the UMLS manual.

\subsection{Key Concepts in the UMLS Metathesaurus}
Let $T$ = ($S_{STR}$, $S_{SRC}$, $S_{SCUI}$, $S_{SG}$) be the set of all input tuples in the Metathesaurus where $S_{STR}$ is the set of all strings, $S_{SRC}$ is the set of all sources, $S_{SCUI}$ is the set of all source concept unique identifiers, and $S_{SG}$ is the set of all semantic groups.

\textbf{AUI and $m_a$ link mapping function.} Every occurrence of a term in a source vocabulary is assigned a unique atom identifier (AUI). 
Let $S_{AUI}$ be the set of all AUIs in the Metathesaurus. Let $m_a$ be the function that maps concept string $str \in S_{STR}$ from source vocabulary $src \in S_{SRC}$ to a new AUI $a \in S_{AUI}$ such that $a$ = $m_a(str, src)$.

\textbf{SUI and $m_s$.} These AUIs are then linked to a unique string identifier (SUI) to represent occurrences of the same string. Any lexical variation in character set, upper-lower case, or punctuation will result in a separate SUI. 
Let $S_{SUI}$ be the set of all SUIs in the Metathesaurus. Let $m_s$ be the function that maps an AUI $a \in S_{AUI}$ to a new SUI $s \in S_{SUI}$ such that $s$ = $m_s(a)$.

\textbf{LUI and $m_l$.} 
All the English lexical variants of a given string (detected using the Lexical Variant Generator tool \cite{mccray1994lexical}) are associated with a single normalized term (LUI). 
Let $S_{LUI}$ be the set of all LUIs in the Metathesaurus. Let $m_l$ be the function that maps a SUI $s \in S_{SUI}$ to a new LUI $l \in S_{LUI}$ such that $l$ = $m_l(s)$.

\textbf{SCUI and $m_s$}. Each AUI is optionally associated with one identifier provided by its source (``Source CUI'' or SCUI). 
Let $S_{SCUI}$ be the set of all SCUIs in the Metathesaurus. Let $m_u$ be the function that maps a concept string $a \in S_{AUI}$ to a new SCUI $u \in S_{SCUI}$ such that $u$ = $m_u(a)$.

\textbf{CUI.} Lexical similarity forms the basis for suggesting synonymy in the UMLS Metathesaurus.
Let $S_{CUI}$ be the set of all concepts CUIs.

\textbf{Source Semantic Group and $m_g$.}  We obtain the semantic group from the semantic type of the concept CUI. Let $S_{SG}$ be the set of semantic groups in the Metathesaurus. 
Let $m_g$ be the function that maps an SCUI $s \in S_{SCUI}$ to a set of semantic groups such that $m_g(s) \subset S_{SG}$.

\textbf{Source Hierarchical Relations and $m_h$.} An SCUI may have parent or child terms in a source vocabulary. 
Let $m_h$ be the function that maps an SCUI $s \in S_{SCUI}$ to a set of its parents, $m_h: S_{SCUI} \rightarrow S_{SCUI}$ such that $m_h(s) \subset S_{SCUI}$.

We will use these key concepts in the remaining sections.

\subsection{The UVA Problem}
The synonymy prediction task is defined in \cite{nguyen2021biomedical} as follows.

$T$ is the set of all input tuples ($S_{STR}$, $S_{SRC}$, $S_{SCUI}$, $S_{SG}$) from source vocabularies.
Let $t$ = ($str$, $src$, $scui$, $sg$) $\in$ $T$, and $t'$ = ($str'$, $src'$, $scui'$, $sg'$) $\in$ $T$.
Let $p$: $T$ $\times$ $T$ $\rightarrow$ \{0,1\} be the prediction function mapping a pair of input tuples to either 0 or 1.
The two strings $str$ from $t$ and $str'$ from $t'$ are synonymous if $p$($t$,$t'$) = 1.

\section{UVA Dataset Generator}
This section summarizes the key points about the UVA data generator implemented in \cite{nguyen2021biomedical} and the datasets generated by this generator. Additional details can be found in this referenced paper.

\label{generator}
\textbf{Ground truth.} Labeled data are taken from the pairs of atoms that are linked to the same (positive) or different (negative) concepts.
Let $POS$ be the set of positive pairs and $NEG$ be the set of negative pairs.

Given a pair of tuples $t$ = ($str$, $src$, $scui$, $sg$) and $t'$ = ($str'$, $src'$, $scui'$, $sg'$), $a$ = $m_a$($str$, $src$), $a'$ = $m_a$($str'$, $src'$), let $m_c$ be the mapping function respectively linking $a, a' \in S_{AUI}$ to $c, c' \in S_{CUI}$ such that $c$ = $m_c$($a$) and $c'$ = $m_c$($a'$), if $c$ = $c'$ then ($a$, $a'$) $\in POS$ else ($a$, $a'$) $\in NEG$.

\textbf{Data generation principles.} Two principles are used to generate the experimental datasets: (1) provide different degrees of lexical similarity in the negative pairs, and (2) maximize the coverage of AUIs in the training datasets. 

Jaccard index is used as a measure for the similarity between atoms. To ignore minor variation among atoms (e.g., singular/plural differences), the lexical similarity of normalized strings is used rather than original strings. 
Let $norm$ be the normalizing function that maps a $sui$ to its normalized string, and $m_s$ be the function mapping an AUI to its SUI. The JACC score assessing the similarity between two AUIs is computed as follows.
\begin{equation}
JACC(a, a') = \dfrac{| norm(m_s(a))\cap norm(m_s(a'))|}{|norm(m_s(a)) \cup norm(m_s(a'))|} 
\end{equation}

\textbf{Degrees of similarity in negative pairs.} All of the negative pairs in the Metathesaurus are divided into two mutually exclusive sets: (1) SIM, the negative pairs with some similarity ($JACC > 0$) between the two atoms, and (2) NOSIM, the negative pairs that have no similarity ($JACC = 0$) between the two atoms. 

Given a pair of tuples $t$ = ($str$, $src$, $scui$, $sg$) and $t'$ = ($str'$, $src'$, $scui'$, $sg'$), with $a$ = $m_a$($str$, $src$), $a'$ = $m_a$($str'$, $src'$), and ($str$, $str'$) $\in NEG$, if $JACC(a, a') > 0$, then ($a, a'$) $\in$ SIM, else ($a, a'$) $\in$ NOSIM.

\textbf{Variants of the negative dataset.} Using the two principles described above, four variants of the negative dataset are created as follows.\\
$NEG_{TOPN}(SIM)$: negative pairs with the highest similarity scores.\\
$NEG_{RAN}(SIM)$: random negative pairs having some similarity.\\
$NEG_{RAN}(NOSIM)$: random negative pairs having no similarity.\\
$NEG_{ALL}$ = $NEG_{TOPN}(SIM)$ $\cup$ $NEG_{RAN}(SIM)$ $\cup$ $NEG_{RAN}(NOSIM)$: include all of the above pairs.

Formally, the number of positive and negative pairs in each dataset variant is computed as follows.
Let $m_c$ be the ground truth function mapping an AUI $a$ to its concept CUI $c$, $c$ = $m_c(a)$.
Let $m_{ca}$ be the function mapping a CUI $c$ to its AUIs $a$, then $m_{ca}(c)$ = \{$a: c = m_c(a)$\}.
Let $n$ be the number of AUIs within a CUI, then $n(a) = |m_{ca}(m_c(a))|$ = |\{$a': c = m_c(a')$\}|.
Let ($a, a'$) be an ordered pair of AUIs, then for every AUI $a$ having $k = (n(a) - 1)$ positive pairs (a,*).
$NEG_{TOPN}(SIM)$ includes $2*k$ negative pairs (a,*) or only 1 negative pair if k = 0.
$NEG_{RAN}(SIM)$ includes $2*k$ negative pairs (a,*) or only 1 negative pair if k = 0.
$NEG_{RAN}(NOSIM)$ includes $2*k$ negative pairs (a,*).
$NEG_{ALL}$ includes up to $6*k$ negative pairs (a,*).

Two types of datasets: (1) learning datasets for training and validating the neural network models, and (2) generalization datasets for testing the generalization of the neural network models. The datasets of the two types are mutually exclusive.

Four dataset variants (TOPN\_SIM, RAN\_SIM, RAN\_NOSIM, and ALL) are created for each dataset type. The set of positive pairs, POS, is split randomly into the learning and generalization datasets (80:20 ratio). The positive learning datasets (80\% of POS) will be combined with the one half of the negative dataset for a given variant. Similarly, the positive generalization datasets (20\% of POS) will be combined with other half of the negative datasets for a given variant.

\section{UVA Baselines}
\label{baselines}

This section summarizes the main points from the three existing baselines from our prior work including the rule-based approximation, the LexLMs \cite{nguyen2021biomedical}, and the ConLMs \cite{nguyen2022conlm}. Additional details can be found in these original papers.

\subsection{Rule-based Approximation (RBA)}

Our RBA baseline \cite{nguyen2021biomedical} uses a set of editorial rules that approximates the current Metathesaurus construction process.

\textbf{Source synonymy (SS) rule.} The two input tuples are synonymous if they have the same identifier in a given source (SCUI). Formally, given a tuple pair $t$ = ($str$, $src$, $scui$, $sg$) $\in T$ and $t'$ = ($str'$, $src'$, $scui'$, $sg'$) $\in T$, let $p_{ss}$ be the prediction function for the source synonymy rule: if $scui$ = $scui'$ then $p_{ss}$($t$, $t'$) = 1. 

\textbf{Lexical similarity and semantic compatibility (LS\_SC) rule.} Given a tuple pair $t$ = ($str$, $src$, $scui$, $sg$) $\in T$ and $t'$ = ($str'$, $src'$, $scui'$, $sg'$) $\in T$, let $p_{lssc}$ be the prediction function for the lexical and semantic similarity rule: $p_{lssc}$($t$,$t'$) = 1 if \\
(1) $aui$ = $m_a(str, src)$, $sui$ = $m_s(aui)$, $lui$ = $m_l(sui)$, $sg$ = $m_g(aui)$,               (deriving lui and sg)
$aui'$ = $m_a(str', src')$, $sui'$ = $m_s(aui')$, $lui'$ = $m_l(sui')$, $sg'$ = $m_g(aui')$,\\
(2) $lui$ = $lui'$ and $sg$ $\cap$ $sg'$ $\neq$ $\emptyset$ (asserting lui and sg).

 \textbf{Rule combination (SS\_LS\_SC)}.
Given a tuple pair $t$ = ($str$, $src$, $scui$, $sg$) $\in T$ and $t'$ = ($str'$, $src'$, $scui'$, $sg'$) $\in T$, let $p_{sslssc}$ be the prediction function for the source synonymy and the lexical and semantic similarity rule: $p_{sslssc}$($t$,$t'$) = 1 if $p\_{ss}$($t$, $t'$) = 1 or $p\_{lssc}$($t$, $t'$) = 1. 

  \textbf{Transitivity.} The combination rule $SS\_LS\_SC$ can be further amplified by considering its transitive closure. 
  Given $t_1, t_2, t_3$ $\in T$, let $p_{trans}$ be the prediction function for the transitivity rule: $P$ = \{$p_{ss}$, $p_{lssc}$, $p_{sslssc}$, $p_{trans}$\} is the set of prediction functions, $p_{trans}$($t_1$, $t_3$) = 1 if $\exists$ $p_1, p_2 \in P$ such that $p_1$($t_1$, $t_2$) = 1 and $p_2$($t_2$, $t_3$) = 1. 

\begin{figure}[t]
\centering
    \includegraphics[width=.99\textwidth]{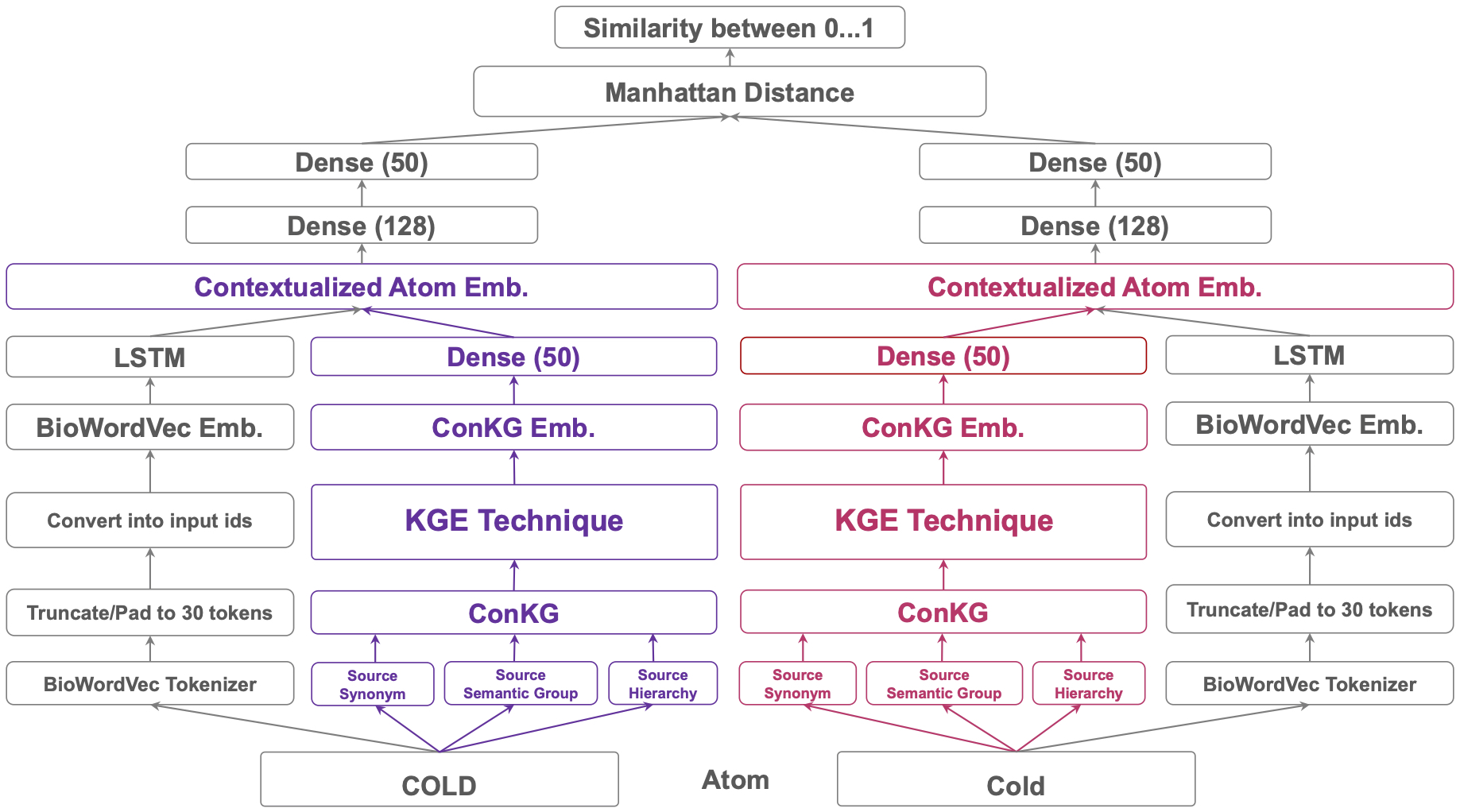}
    \caption{The architecture of the neural network in the Context-enriched Learning Model (ConLM) \cite{nguyen2022conlm} created by adding the ConKG embeddings to the Lexical-based Learning Model (LexLM) embeddings (grey boxes) from \cite{nguyen2021biomedical}. The seven KGE technique used for ConKG include TransE, TransR, DistMult, HolE, ComplEx, RESCAL, and ConvKB. }
    \label{fig:nn}
\end{figure}

\subsection{LexLMs} 
The LexLM \cite{nguyen2021biomedical} (grey boxes in Figure \ref{fig:nn}) adopts the Siamese architecture \cite{mueller2016siamese} that takes in a pair of inputs and learns representations based on explicit similarity and dissimilarity information defined during training. The inputs (a pair of atoms) are pre-processed and transformed into their numerical representations with BioWordVec embeddings pre-trained from PubMed text corpus and MeSH data \cite{zhang2019biowordvec}. The word embeddings are then fed to Long Short Term Memory (LSTM) layers to learn the semantic and syntactic features of the atoms through time. The LexLM relies exclusively on the lexical features of the atoms, i.e., the terms themselves.

\subsection{ConLMs}
\renewcommand{\arraystretch}{1}
\begin{table}
\centering
\caption{Set of ConKG embeddings for each ConKG variant}
\label{tab:context_derive} 
\resizebox{0.99\textwidth}{!}{%
\begin{tabular}{@{}|c|l|l|@{}}
\toprule
\multicolumn{1}{c|}{\textbf{Variant}} &
\multicolumn{1}{c|}{\textbf{ConKG Triples}} &
\multicolumn{1}{c|}{\textbf{Set of ConKG embedding vectors}} \\ \midrule
$\mathbf{ConSS}$     & $\{(a, r_s, s): s = m_s(a)\}$ & $\{\mathbf{E}_{ConSS}(a) \oplus \mathbf{E}_{ConSS}(m_s(a)): \forall a \in S_{AUI}\}$           \\ \midrule
$\mathbf{ConSG}$    & $\{(s, r_g, g): g \in m_g(s)\}$ & $\{\mathbf{E}_{ConSG}(m_s(a)) \oplus  \avsum_{j=1}^{\|m_g(m_s(a))\|} \mathbf{E}_{ConSG}(g_j): \forall a \in S_{AUI}, g_j \in m_g(m_s(a))\}$          \\ \midrule
$\mathbf{ConHR}$   & $\{(s, r_h, p): p \in m_h(s)\}$ & $\{\mathbf{E}_{ConHR}(m_s(a)): \forall a \in S_{AUI}\}$          \\ \midrule
\multirow{3}{*}{$\mathbf{ConAll}$}   & $\{(a, r_s, s): s = m_s(a)\}$  &  $\{\mathbf{E}_{ConAll}(a) \oplus$ \\
 & $\{(s, r_g, g): g \in m_g(s)\}$ &  $\mathbf{E}_{ConAll}(m_s(a)) \oplus$ \\
 & $\{(s, r_h, p): p \in m_h(s)\}$ & $\avsum_{j=1}^{\|m_g(m_s(a))\|} \mathbf{E}_{ConAll}(g_j) : \forall a \in S_{AUI}, g_j \in m_g(m_s(a))\}$ \\ \bottomrule
\end{tabular}%
}
\end{table}

The ConLMs \cite{nguyen2022conlm} extends the LexLMs by incorporating the contextual information such as semantic groups, source synonymy, and source hierarchy into the neural networks. The contextual information is represented in the ConKGs as summarized in Table \ref{tab:context_derive}.

$S_{AUI}$ is the set of AUIs, $S_{SCUI}$ is the set of SCUIs, $S_{SG}$ is the set of SGs, and the mapping functions \{$m_s, m_g, m_h$\} are defined in Section \ref{background} above.

\textbf{ConSS.} Let $r_s$ denote the binary relation has\_SCUI from an AUI $a \in S_{AUI}$ to an SCUI $s \in S_{SCUI}$. The ConSS variant includes the triples representing the relationship between an AUI and its SCUI.

\textbf{ConSG.} Let $r_g$ denote the binary relation has\_SG from an SCUI $s \in S_{SCUI}$ to a SG $g \in S_{SG}$. The ConSG variant includes the triples representing the relationship between an SCUI and its semantic groups.

\textbf{ConHR.} Let $r_h$ denote the binary relation has\_parentSCUI from an SCUI $s \in S_{SCUI}$ to its parent SCUI $p \in S_{SCUI}$. The ConHR variant includes the triples representing the relationship between an SCUI and its parent SCUI.

\textbf{ConAll}. ConAll = ConSS $\cup$ ConSG $\cup$ ConHR. 

The ConKG triples can be transformed into their knowledge graph embedding vectors using various KG embedding techniques such as TransE \cite{bordes2013translating}, TransR \cite{lin2015learning}, RESCAL \cite{nickel2011three}, DistMult \cite{yang2014embedding}, HolE \cite{nickel2016holographic}, ComplEx \cite{trouillon2016complex}, and ConvKB \cite{nguyen2017novel}

Let $E$ be the set of ConKG entities, $E = S_{AUI} \cup S_{SCUI} \cup S_{SG}$.
Let $R$ be the set of all ConKG relations, $R = \{r_s, r_g, r_h\}$.
Let $\mathbf{T}$ be the set of ConKG triples, a triple $t \in \mathbf{T}$ if $t = (e_1, r, e_2)$ with $r \in R$, and $e_1,e_2 \in E$.
Let $\mathbf{T'}$ be the set of negative ConKG triples, $t' = (e'_1, r, e'_2) \in \mathbf{T'}$ if $\exists t = (e_1, r, e_2) \in \mathbf{T}$ and $t' \notin \mathbf{T} $. Let $d = 2*i$ ($i \in \mathbf{N}$) be the dimension of embedding vector. We choose $d$ to be an even number to facilitate the representation of ComplEx vectors.

\textbf{Entity embeddings.} Let $\mathbf{E}$ be the set of embedding vectors of $E$, then $\mathbf{e} \in \mathbf{E}$ is an embedding vector of entity $e$. 
While the embedding for an entity or relation from TransE, HolE, and DistMult is a single vector with dimension $d = 2i$, ComplEx embeddings require two vectors (real and imaginary) of dimension $d = i$. In this case, we concatenate the real and imaginary vectors for each entity into a single vector of dimension $d = 2i$. For ComplEx, we denote the two-vector embeddings as $\textbf{E} = (\textbf{E}_{re}$, $\textbf{E}_{im})$, then we define the embedding vector for an entity $e$ as $\textbf{E}(e) = \textbf{E}_{re}(e) \oplus \textbf{E}_{im}(e)$, $\forall e \in \textbf{E}$.
 
The output of each KG embedding technique is a set of entity embeddings: $\mathbf{E}_{ConSS}$, $\mathbf{E}_{ConSG}$, $\mathbf{E}_{ConHR}$, and $\mathbf{E}_{ConAll}$, which will be used to derive the ConKG embeddings in Table \ref{tab:context_derive}.

The ConLM \cite{nguyen2022conlm} (Figure \ref{fig:nn}) adds to the LexLM (at the LSTM layer) a specific variant of the ConKGs described in Table \ref{tab:context_derive}. For each ConKG variant, the respective trained ConKG embedding vectors are fed to a 50-unit dense layer to learn the derived features in Table \ref{tab:context_derive}.  The output of the dense layer is then concatenated ($\oplus$)  with the output of the LSTM units from the LexLM. Together, they form the contextualized atom embedding, which is then fed to subsequent dense layers with 128 and 50 learning units, respectively. The output is a Manhattan distance similarity function \cite{aggarwal2001surprising}, which computes a score indicating the degree of synonymy between the atoms with a threshold of 0.5. 

\section{Experiments}
\label{experiments}

This section presents the set of experiments for demonstrating the reusability and reproducibility of the UVA baselines described in Section \ref{baselines}. We describe the dataset generation in Section \ref{generating_datasets} and the baseline experiments in Section \ref{generating_baselines}.

\subsection{Experimental Setup}
All these experiments are deployed as batches of parallel jobs with the Slurm~\footnote{https://slurm.schedmd.com/documentation.html} workload manager to the Biowulf high-performance computing cluster~\footnote{https://hpc.nih.gov/} at the National Institutes of Health (NIH). 
The \textit{norm} and \textit{gpu} partitions with their corresponding CPU and GPU servers in this cluster (with a limit of 10,000 CPU cores, 60 TB of RAM, and 56 GPUs per user) are used. The following settings are used for deployment to maximize the resources allocated in Biowulf to reduce runtime: (1) multiple nodes (500-625 nodes), (2) multiple threadings with 16-20 threads per node, (3) approximately 95-125 GB of RAM per node.
We used Tesla V100x GPUs with 32GB of GPU RAM and at least 220GB of CPU RAM for each training and testing task.
While the implementation is configurable and reproducible in a different environment, these experiments are computationally- and resource-intensive. 
\begin{table}[h]
\centering
\caption{The statistics of UVA datasets generated from UMLS releases 2020AA, 2021AA, and 2021AB \label{statistics}}
\begin{tabular}{@{}|r|r|r|r|@{}}
\toprule
\textbf{Dataset}            & \multicolumn{1}{r|}{\textbf{NEG}} & \multicolumn{1}{r|}{\textbf{POS}} & \multicolumn{1}{r|}{\textbf{Total}} \\ \midrule
20AA\_TRAIN\_ALL                     & 101,322,647                           & 16,743,627                            & 118,066,274                        \\ \midrule
{\color[HTML]{548235} 20AA\_GEN\_ALL}                        & {\color[HTML]{548235} 166,410,710}   & {\color[HTML]{548235} 5,581,208}    & {\color[HTML]{548235} 171,991,918}                        \\ \midrule
{\color[HTML]{305496} 20AA\_GEN\_TOPN\_SIM}                 & {\color[HTML]{305496} 54,752,228}     & {\color[HTML]{305496} 5,581,208}   & {\color[HTML]{305496} 60,333,436}                         \\ \midrule
{\color[HTML]{7030A0} 20AA\_GEN\_RAN\_SIM}                   & {\color[HTML]{7030A0} 54,445,899}    & {\color[HTML]{7030A0} 5,581,208}  & {\color[HTML]{7030A0} 60,027,107}                         \\ \midrule
{\color[HTML]{203764} 20AA\_GEN\_RAN\_NOSIM}                 & {\color[HTML]{203764} 58,256,526}    & {\color[HTML]{203764} 5,581,208}  & {\color[HTML]{203764} 63,837,734}                         \\ \midrule
\textbf{Dataset}                            & \multicolumn{1}{r|}{\textbf{NEG}}  & \multicolumn{1}{r|}{\textbf{POS}} & \multicolumn{1}{r|}{\textbf{Total}} \\ \midrule
21AA\_TRAIN\_ALL                            & 133,165,966                        & 22,614,004                        & 155,779,960                         \\ \midrule
{\color[HTML]{548235} 21AA\_GEN\_ALL}       & {\color[HTML]{548235} 219,495,843} & {\color[HTML]{548235} 7,537,999}  & {\color[HTML]{548235} 227,033,842}  \\ \midrule
{\color[HTML]{305496} 21AA\_GEN\_TOPN\_SIM} & {\color[HTML]{305496} 72,286,689}  & {\color[HTML]{305496} 7,537,999}  & {\color[HTML]{305496} 79,824,688}   \\ \midrule
{\color[HTML]{7030A0} 21AA\_GEN\_RAN\_SIM}  & {\color[HTML]{7030A0} 72,124,757}  & {\color[HTML]{7030A0} 7,537,999}  & {\color[HTML]{7030A0} 79,662,756}   \\ \midrule
21AA\_GEN\_RAN\_NOSIM                       & 76,308,150                         & 7,537,999                         & 83,846,149                          \\ \midrule
\textbf{Dataset}                            & \multicolumn{1}{r|}{\textbf{NEG}}  & \multicolumn{1}{r|}{\textbf{POS}} & \multicolumn{1}{r|}{\textbf{Total}} \\ \midrule
21AB\_TRAIN\_ALL                            & 133,663,761                        & 22,438,339                        & 156,102,100                         \\ \midrule
{\color[HTML]{548235} 21AB\_GEN\_ALL}       & {\color[HTML]{548235} 220,138,372} & {\color[HTML]{548235} 7,479,445}  & {\color[HTML]{548235} 227,617,817}  \\ \midrule
{\color[HTML]{305496} 21AB\_GEN\_TOPN\_SIM} & {\color[HTML]{305496} 72,646,487}  & {\color[HTML]{305496} 7,479,445}  & {\color[HTML]{305496} 80,125,932}   \\ \midrule
{\color[HTML]{7030A0} 21AB\_GEN\_RAN\_SIM}  & {\color[HTML]{7030A0} 72,319,075}  & {\color[HTML]{7030A0} 7,479,445}  & {\color[HTML]{7030A0} 79,798,250}   \\ \midrule
21AB\_GEN\_RAN\_NOSIM                       & 76,413,759                         & 7,479,445                         & 83,893,204                          \\ \bottomrule
\end{tabular}
\end{table}

\subsection{Generating UVA Datasets}
\label{generating_datasets}

\begin{table}[h]
\centering
\caption{Experimental results for the two UVA baselines (ConLM and LexLM) using the 2020AA-ACTIVE datasets with different levels of lexical similarity\label{2020AA-results}}
\begin{tabular}{@{}|rrrrrrrrr|@{}}
\toprule
\multicolumn{9}{|c|}{\textbf{2020AA-ACTIVE}}                                                                                                                                                                                                                                                                                                                                                                                                                \\ \midrule
\multicolumn{1}{|r|}{}                           & \multicolumn{4}{c|}{{\color[HTML]{548235} \textbf{GEN\_ALL}}}                                                                                                                                                 & \multicolumn{4}{c|}{{\color[HTML]{305496} \textbf{GEN\_TOPN\_SIM}}}                                                                                                                      \\ \cmidrule(l){2-9} 
\multicolumn{1}{|r|}{\multirow{-2}{*}{Baseline}} & \multicolumn{1}{r|}{accuracy}                    & \multicolumn{1}{r|}{precision}                    & \multicolumn{1}{r|}{recall}                       & \multicolumn{1}{r|}{F1}                           & \multicolumn{1}{r|}{accuracy}                    & \multicolumn{1}{r|}{precision}                    & \multicolumn{1}{r|}{recall}                       & F1                           \\ \midrule
\multicolumn{1}{|r|}{RBA}                        & \multicolumn{1}{r|}{{\color[HTML]{548235} 98.63}} & \multicolumn{1}{r|}{{\color[HTML]{548235} 86.31}} & \multicolumn{1}{r|}{{\color[HTML]{548235} 68.71}} & \multicolumn{1}{r|}{{\color[HTML]{548235} 76.51}} & \multicolumn{1}{r|}{{\color[HTML]{2F75B5} 96.14}} & \multicolumn{1}{r|}{{\color[HTML]{2F75B5} 86.83}} & \multicolumn{1}{r|}{{\color[HTML]{2F75B5} 68.71}} & {\color[HTML]{2F75B5} 76.72} \\ \midrule
\multicolumn{1}{|r|}{LexLM}                      & \multicolumn{1}{r|}{{\color[HTML]{548235} 99.38}} & \multicolumn{1}{r|}{{\color[HTML]{548235} 88.75}} & \multicolumn{1}{r|}{{\color[HTML]{548235} 92.54}} & \multicolumn{1}{r|}{{\color[HTML]{548235} 90.61}}  & \multicolumn{1}{r|}{{\color[HTML]{2F75B5} 98.07}} & \multicolumn{1}{r|}{{\color[HTML]{2F75B5} 88.41}} & \multicolumn{1}{r|}{{\color[HTML]{2F75B5} 91.10}} & {\color[HTML]{2F75B5} 89.74} \\ \midrule
\multicolumn{1}{|r|}{ConLM}                      & \multicolumn{1}{r|}{{\color[HTML]{548235} 99.58}} & \multicolumn{1}{r|}{{\color[HTML]{548235} 93.75}} & \multicolumn{1}{r|}{{\color[HTML]{548235} 93.23}} & \multicolumn{1}{r|}{{\color[HTML]{548235} 93.49}} & \multicolumn{1}{r|}{{\color[HTML]{2F75B5} 98.92}} & \multicolumn{1}{r|}{{\color[HTML]{2F75B5} 94.97}} & \multicolumn{1}{r|}{{\color[HTML]{2F75B5} 93.23}} & {\color[HTML]{2F75B5} 94.09} \\ \midrule
\multicolumn{1}{|r|}{}                           & \multicolumn{4}{c|}{{\color[HTML]{7030A0} \textbf{GEN\_RAN\_SIM}}}                                                                                                                                            & \multicolumn{4}{c|}{\textbf{GEN\_RAN\_NOSIM}}                                                                                                                                            \\ \cmidrule(l){2-9} 
\multicolumn{1}{|r|}{\multirow{-2}{*}{Baseline}} & \multicolumn{1}{r|}{accuracy}                    & \multicolumn{1}{r|}{precision}                    & \multicolumn{1}{r|}{recall}                       & \multicolumn{1}{r|}{F1}                           & \multicolumn{1}{r|}{accuracy}                    & \multicolumn{1}{r|}{precision}                    & \multicolumn{1}{r|}{recall}                       & F1                           \\ \midrule
\multicolumn{1}{|r|}{RBA}                        & \multicolumn{1}{r|}{{\color[HTML]{7030A0} 97.02}} & \multicolumn{1}{r|}{{\color[HTML]{7030A0} 98.92}} & \multicolumn{1}{r|}{{\color[HTML]{7030A0} 68.71}} & \multicolumn{1}{r|}{{\color[HTML]{7030A0} 81.09}} & \multicolumn{1}{r|}{{\color[HTML]{203764} 97.26}} & \multicolumn{1}{r|}{{\color[HTML]{203764} 99.99}} & \multicolumn{1}{r|}{{\color[HTML]{203764} 68.71}} & {\color[HTML]{203764} 81.45} \\ \midrule
\multicolumn{1}{|r|}{LexLM}                      & \multicolumn{1}{r|}{{\color[HTML]{7030A0} 99.05}} & \multicolumn{1}{r|}{{\color[HTML]{7030A0} 98.58}} & \multicolumn{1}{r|}{{\color[HTML]{7030A0} 91.10}} & \multicolumn{1}{r|}{{\color[HTML]{7030A0} 94.69}} & \multicolumn{1}{r|}{{\color[HTML]{203764} 99.24}} & \multicolumn{1}{r|}{{\color[HTML]{203764} 99.71}} & \multicolumn{1}{r|}{{\color[HTML]{203764} 91.62}} & {\color[HTML]{203764} 95.49} \\ \midrule
\multicolumn{1}{|r|}{ConLM}                      & \multicolumn{1}{r|}{{\color[HTML]{7030A0} 99.28}} & \multicolumn{1}{r|}{{\color[HTML]{7030A0} 98.54}} & \multicolumn{1}{r|}{{\color[HTML]{7030A0} 93.23}} & \multicolumn{1}{r|}{{\color[HTML]{7030A0} 96.00}} & \multicolumn{1}{r|}{{\color[HTML]{203764} 99.38}} & \multicolumn{1}{r|}{{\color[HTML]{203764} 99.67}} & \multicolumn{1}{r|}{{\color[HTML]{203764} 93.23}} & {\color[HTML]{203764} 96.34} \\ \bottomrule
\end{tabular}
\end{table}

\begin{table}[h]
\centering
\caption{Experimental results for the two UVA baselines (ConLM and LexLM) using the 2021AA-ACTIVE datasets with different levels of lexical similarity\label{2021AA-results}}
\begin{tabular}{@{}|rrrrrrrrr|@{}}
\toprule
\multicolumn{9}{|c|}{\textbf{2021AA-ACTIVE}}                                                                                                                                                                                                                                                                                                                                                                                                                \\ \midrule
\multicolumn{1}{|c|}{}                           & \multicolumn{4}{c|}{{\color[HTML]{548235} \textbf{GEN\_ALL}}}                                                                                                                                                 & \multicolumn{4}{c|}{{\color[HTML]{305496} \textbf{GEN\_TOPN\_SIM}}}                                                                                                                      \\ \cmidrule(l){2-9} 
\multicolumn{1}{|c|}{\multirow{-2}{*}{Baseline}} & \multicolumn{1}{c|}{accuracy}                    & \multicolumn{1}{c|}{precision}                    & \multicolumn{1}{c|}{recall}                       & \multicolumn{1}{c|}{F1}                           & \multicolumn{1}{c|}{accuracy}                    & \multicolumn{1}{c|}{precision}                    & \multicolumn{1}{c|}{recall}                       & \multicolumn{1}{c|}{F1}      \\ \midrule
\multicolumn{1}{|r|}{LexLM}                      & \multicolumn{1}{r|}{{\color[HTML]{548235} 99.16}} & \multicolumn{1}{r|}{{\color[HTML]{548235} 87.14}} & \multicolumn{1}{r|}{{\color[HTML]{548235} 87.74}} & \multicolumn{1}{r|}{{\color[HTML]{548235} 87.44}} & \multicolumn{1}{r|}{{\color[HTML]{2F75B5} 97.74}} & \multicolumn{1}{r|}{{\color[HTML]{2F75B5} 88.22}} & \multicolumn{1}{r|}{{\color[HTML]{2F75B5} 87.74}} & {\color[HTML]{2F75B5} 87.98} \\ \midrule
\multicolumn{1}{|r|}{ConLM}                      & \multicolumn{1}{r|}{{\color[HTML]{548235} 99.51}} & \multicolumn{1}{r|}{{\color[HTML]{548235} 91.43}} & \multicolumn{1}{r|}{{\color[HTML]{548235} 93.96}} & \multicolumn{1}{r|}{{\color[HTML]{548235} 92.98}} & \multicolumn{1}{r|}{{\color[HTML]{2F75B5} 98.80}} & \multicolumn{1}{r|}{{\color[HTML]{2F75B5} 93.41}} & \multicolumn{1}{r|}{{\color[HTML]{2F75B5} 93.96}} & {\color[HTML]{2F75B5} 93.69} \\ \midrule
\multicolumn{1}{|c|}{}                           & \multicolumn{4}{c|}{{\color[HTML]{7030A0} \textbf{GEN\_RAN\_SIM}}}                                                                                                                                            & \multicolumn{4}{c|}{\textbf{GEN\_RAN\_NOSIM}}                                                                                                                                            \\ \cmidrule(l){2-9} 
\multicolumn{1}{|c|}{\multirow{-2}{*}{Baseline}} & \multicolumn{1}{c|}{accuracy}                    & \multicolumn{1}{c|}{precision}                    & \multicolumn{1}{c|}{recall}                       & \multicolumn{1}{c|}{F1}                           & \multicolumn{1}{c|}{accuracy}                    & \multicolumn{1}{c|}{precision}                    & \multicolumn{1}{c|}{recall}                       & \multicolumn{1}{c|}{F1}      \\ \midrule
\multicolumn{1}{|r|}{LexLM}                      & \multicolumn{1}{r|}{{\color[HTML]{7030A0} 98.73}} & \multicolumn{1}{r|}{{\color[HTML]{7030A0} 98.73}} & \multicolumn{1}{r|}{{\color[HTML]{7030A0} 87.74}} & \multicolumn{1}{r|}{{\color[HTML]{7030A0} 92.91}} & \multicolumn{1}{r|}{{\color[HTML]{203764} 98.87}} & \multicolumn{1}{r|}{{\color[HTML]{203764} 99.69}} & \multicolumn{1}{r|}{{\color[HTML]{203764} 87.74}} & {\color[HTML]{203764} 98.33} \\ \midrule
\multicolumn{1}{|r|}{ConLM}                      & \multicolumn{1}{r|}{{\color[HTML]{7030A0} 99.28}} & \multicolumn{1}{r|}{{\color[HTML]{7030A0} 98.34}} & \multicolumn{1}{r|}{{\color[HTML]{7030A0} 93.96}} & \multicolumn{1}{r|}{{\color[HTML]{7030A0} 96.10}} & \multicolumn{1}{r|}{{\color[HTML]{203764} 99.40}} & \multicolumn{1}{r|}{{\color[HTML]{203764} 99.32}} & \multicolumn{1}{r|}{{\color[HTML]{203764} 93.96}} & {\color[HTML]{203764} 96.57} \\ \bottomrule
\end{tabular}
\end{table}
\begin{table}[h]
\centering
\caption{Experimental results for the two UVA baselines (ConLM and LexLM) using the 2021AB-ACTIVE datasets with different levels of lexical similarity\label{2021AB-results}}
\begin{tabular}{@{}|rrrrrrrrr|@{}}
\toprule
\multicolumn{9}{|c|}{\textbf{2021AB-ACTIVE}}                                                                                                                                                                                                                                                                                                                                                                                                                         \\ \midrule
\multicolumn{1}{|c|}{}                           & \multicolumn{4}{c|}{{\color[HTML]{548235} \textbf{GEN\_ALL}}}                                                                                                                                                 & \multicolumn{4}{c|}{{\color[HTML]{305496} \textbf{GEN\_TOPN\_SIM}}}                                                                                                                               \\ \cmidrule(l){2-9} 
\multicolumn{1}{|c|}{\multirow{-2}{*}{Baseline}} & \multicolumn{1}{c|}{accuracy}                    & \multicolumn{1}{c|}{precision}                    & \multicolumn{1}{c|}{recall}                       & \multicolumn{1}{c|}{F1}                           & \multicolumn{1}{c|}{accuracy}                    & \multicolumn{1}{c|}{precision}                    & \multicolumn{1}{c|}{recall}                       & \multicolumn{1}{c|}{F1}               \\ \midrule
\multicolumn{1}{|r|}{LexLM}                      & \multicolumn{1}{r|}{{\color[HTML]{548235} 99.37}} & \multicolumn{1}{r|}{{\color[HTML]{548235} 88.18}} & \multicolumn{1}{r|}{{\color[HTML]{548235} 93.25}} & \multicolumn{1}{r|}{{\color[HTML]{548235} 90.64}} & \multicolumn{1}{r|}{{\color[HTML]{2F75B5} 98.31}} & \multicolumn{1}{r|}{{\color[HTML]{2F75B5} 89.13}} & \multicolumn{1}{r|}{{\color[HTML]{2F75B5} 93.25}} & {\color[HTML]{2F75B5} 91.14}          \\ \midrule
\multicolumn{1}{|r|}{ConLM}                      & \multicolumn{1}{r|}{{\color[HTML]{548235} 99.51}} & \multicolumn{1}{r|}{{\color[HTML]{548235} 91.52}} & \multicolumn{1}{r|}{{\color[HTML]{548235} 93.82}} & \multicolumn{1}{r|}{{\color[HTML]{548235} 92.65}} & \multicolumn{1}{r|}{{\color[HTML]{2F75B5} 98.82}} & \multicolumn{1}{r|}{{\color[HTML]{2F75B5} 93.60}} & \multicolumn{1}{r|}{{\color[HTML]{2F75B5} 93.82}} & {\color[HTML]{2F75B5} \textbf{93.71}} \\ \midrule
\multicolumn{1}{|c|}{}                           & \multicolumn{4}{c|}{{\color[HTML]{7030A0} \textbf{GEN\_RAN\_SIM}}}                                                                                                                                            & \multicolumn{4}{c|}{{\color[HTML]{203764} \textbf{GEN\_RAN\_NOSIM}}}                                                                                                                              \\ \cmidrule(l){2-9} 
\multicolumn{1}{|c|}{\multirow{-2}{*}{Baseline}} & \multicolumn{1}{c|}{accuracy}                    & \multicolumn{1}{c|}{precision}                    & \multicolumn{1}{c|}{recall}                       & \multicolumn{1}{c|}{F1}                           & \multicolumn{1}{c|}{accuracy}                    & \multicolumn{1}{c|}{precision}                    & \multicolumn{1}{c|}{recall}                       & \multicolumn{1}{c|}{F1}               \\ \midrule
\multicolumn{1}{|r|}{LexLM}                      & \multicolumn{1}{r|}{{\color[HTML]{7030A0} 99.27}} & \multicolumn{1}{r|}{{\color[HTML]{7030A0} 98.89}} & \multicolumn{1}{r|}{{\color[HTML]{7030A0} 93.25}} & \multicolumn{1}{r|}{{\color[HTML]{7030A0} 95.98}} & \multicolumn{1}{r|}{99.38}                        & \multicolumn{1}{r|}{99.73}                        & \multicolumn{1}{r|}{93.25}                        & 96.38                                 \\ \midrule
\multicolumn{1}{|r|}{ConLM}                      & \multicolumn{1}{r|}{{\color[HTML]{7030A0} 99.26}} & \multicolumn{1}{r|}{{\color[HTML]{7030A0} 98.18}} & \multicolumn{1}{r|}{{\color[HTML]{7030A0} 93.82}} & \multicolumn{1}{r|}{{\color[HTML]{7030A0} 95.95}} & \multicolumn{1}{r|}{99.40}                        & \multicolumn{1}{r|}{99.38}                        & \multicolumn{1}{r|}{93.82}                        & 96.52                                 \\ \bottomrule
\end{tabular}
\end{table}

We used the dataset generator described in Section \ref{generator} to generate UVA datasets.
We installed three UMLS releases including 2020AA, 2021AA, and 2021AB. We note that the 2022AA release has not been available at the time of this writing. For each UMLS release, we only selected active subset of source vocabularies in English, and the terms not being suppressed.

Table \ref{statistics} shows the statistics of the training and generalization test sets with the number of positive, negative, and total pairs for each UMLS version. These datasets will be used for the experiments described in Section \ref{experiments}.
This generator task required approximately 10,000 CPU cores from 300 to 600 nodes with 90 to 120 GB of RAM per node in the \textit{norm} partition of the Biowulf cluster, but finished within 24 hours.

\subsection{Generating UVA baselines}
\label{generating_baselines}

Here our goal is to demonstrate the reusability and the reproducibility of the baseline resources across multiple UMLS releases. Therefore, we did not perform extensive set of experiments with all variants from each UVA baseline. Instead, we chose the best variant of each baseline from previous experiments \cite{nguyen2022conlm,nguyen2021biomedical} to run the experiments.

For training the LexLMs and ConLMs, we use the TRAIN\_ALL dataset from each UVA version. For testing the overall generalization of each model, we used the GEN\_ALL. For testing the models with different levels of lexical similarity, we used the three corresponding datasets GEN\_TOPN\_SIM, GEN\_RAN\_SIM, and GEN\_RAN\_NOSIM. In the ConLMs, we configured to use ComplEx for training the ConKG embeddings.
We used default hyper parameters for the training and testing of the ConLMs and the LexLMs.

We report the experimental results in Table \ref{2020AA-results}, \ref{2021AA-results}, and \ref{2021AB-results} for each UVA dataset version.
Overall, the performance results of the ConLM, LexLM, and RBA are consistent with the ConLM performing best across the experiments.

\section{Discussion and Future Work}
\label{discussion}
We have presented the UVA resources across multiple UMLS releases to be shared with broader research communities. Here we discuss the directions we could move forward with these resources.

\subsection{Scalability, Reusability, and Reproducibility}

\textbf{Scalability.} Scalability is among the primary challenges at every step in the pipeline of the UVA problem. Every task we have deployed to the NIH's Biowulf HPC would require extensive resources. Especially, the tasks that require pair-wise calculation such as lexical similarity would be computationally heavy. Therefore, optimizing the tasks to run with less resource-extensive requirements within a reasonable timeframe remains a challenge for the project. 

Additionally the scalability in terms of data size also increases when we extend the UVA task for the new applications. For instance, inserting a new vocabulary into the UMLS Metathesaurus would involve pair-wise predictions from the old and new terms.

\textbf{Reusability.} Our resources can be reused in different ways. The dataset generator can be reused for generating new UVA datasets from customized installations of the UMLS. For example, one can choose to install the full set of source vocabularies instead of just selecting the active ones. The data generator can also be extended to generate datasets with new characteristics. For example, we could limit the pairs in the UVA task to be from the same semantic group. Our baselines can be extended for developing the new approaches and even benchmarking them. Indeed, our LexLM has been reused for developing the ConLM. The resulting embeddings from our models can also be used for evaluating other BioNLP tasks such as UMLS entity recognition.

\textbf{Reproducibility.} Our resources presented in this paper are completely reproducible across multiple UMLS releases. We made the source code, the datasets, and other materials available \footnote{https://w3id.org/uva} for download. We also provide the scripts with customized parameters for automating the entire experiments. 







\subsection{UVA Resource Maintenance Plan}
\label{releases}

\textbf{License.}
UMLS license is a no-cost license, required to download any UMLS distribution.
Therefore, the UMLS license is also required for using the UVA resources.

\textbf{UVA Challenge.} We plan to organize an annual challenge inviting new ideas, novel approaches and solutions for the UVA task. We will adopt the UVA resources to generate new datasets and baselines to serve as input for the UVA challenge.

The UVA task differs from the Ontology Alignment Evaluation Initiative (OAEI) in terms of scalability and domain diversity. Particularly, the size of the UVA datasets is several orders of magnitude more than tens of thousands of pairs in the OAEI datasets). Furthermore, the UVA task covers all the biomedical domains in the UMLS while the OAEI only cover a few ontologies. Therefore, we believe that the UVA task will complement the OAEI and offer greater challenges for the community.

\section{Related Work}
\label{related}





Biomedical ontology alignment is a long-standing research effort driven by the Ontology Alignment Evaluation Initiative (OAEI) since 2005. With the growth of interest in integrating biomedical ontologies at scale \cite{nguyen2021biomedical}, studies have looked into using rule-based and statistical approaches \cite{faria2013agreementmakerlight,ngo2016overview,jimenez2011logmap}, as well as supervised learning approaches for ontologies matching \cite{doan2004ontology}  by assessing the similarity \cite{kolyvakis2018biomedical,wang2018ontology} and relatedness \cite{mao2020use} between words and sentences. Such tasks are also known as Semantic Text Similarity (STS) tasks. 

Recent progress has been attributed to the use of a combination of knowledge-based similarity with deep learning techniques, such as word embeddings \cite{mihalcea2006corpus} for input feature representation, and Siamese Network \cite{mueller2016siamese,he2015multi,bento2020ontology,huang2020sentence} to learn the underlying semantics and structure. Unsupervised approaches, such as transformer-based mechanisms, have also been explored with a great degree of success \cite{devlin2018bert}. Nonetheless, these techniques are largely based on lexical features and require very large text corpora to learn from. In \cite{iyer2020veealign}, contextualized representations of concepts with ancestral, child, data property neighbor, and object property neighbor contexts are used to discover semantically equivalent concepts. In our approach, we exploit the graph structure of various types of contextual information through the use of KG embeddings.

There are several families of KG embedding techniques \cite{ji2020survey,sun2020benchmarking}. They aim to map entities and relations into low-dimensional vectors while capturing their structural and semantic meanings \cite{wang2017knowledge}, and have shown to benefit a variety of knowledge-driven tasks \cite{lin2018knowledge}. Since this is the first attempt to use the various types of contextual information in the UVA task at scale, we explored three of the popular classes of techniques because of their demonstrated success in various tasks: translational distance-based with TransE \cite{bordes2013translating}, and TransR \cite{lin2015learning}; semantic matching-based using RESCAL \cite{nickel2011three}, DistMult \cite{yang2014embedding}, HolE \cite{nickel2016holographic}, and ComplEx \cite{trouillon2016complex}; and neural network-based using ConvKB \cite{nguyen2017novel}. We evaluated their performance in the UVA task, but did not benchmark them against other forms of graph representation techniques \cite{ji2020survey,wu2020comprehensive}. Many embedding techniques listed in \cite{sun2020benchmarking} are not selected due to the specific characteristics of context in our datasets, e.g., having no attributes/literals (but 10 techniques in \cite{sun2020benchmarking} including AttrE \cite{trisedya2019entity} and KDCoE \cite{chen2018co} leveraging attributes), having English-only (but MTransE \cite{chen2016multilingual} leveraging multilingual), or very large (RDGCN \cite{wu2019relation} being not scalable).

\section{Conclusion}
\label{conclusion}

We have presented the set of UVA resources including the dataset generator, the datasets, and the baseline approaches to be shared with broader research communities. We have shown how the resources can be reused and reproduced across multiple UMLS releases, and gathered to collectively serve the UVA challenge. 

\paragraph*{Resource Availability Statement:} UVA resources including the source code and the datasets are available from \url{https://w3id.org/uva}.

\paragraph*{Acknowledgement.} This research was supported in part by the Intramural Research Program of the National Library of Medicine (NLM), National Institutes of Health (NIH). This research was also supported in part by two appointments to the National Library of Medicine Research Participation Program. This program is administered by the Oak Ridge Institute for Science and Education through an inter-agency agreement between the U.S. Department of Energy and the National Library of Medicine. We are thankful to Miranda Jarnot for helping us to confirm the editorial rules used in the Metathesaurus. We are also thankful to our summer interns (Hong Yung Yip, Goonmeet Bajaj, Thilini Wijesiriwardene, and Vishesh Javangula) and collaborators (Amit Sheth and Srinivasan Parthasarathy) for contributing to the resources presented in this paper.

\bibliographystyle{abbrv}
\bibliography{main}

\end{document}